\documentclass[sigconf,authordraft]{acmart}
\renewcommand\footnotetextcopyrightpermission[1]{} 
\AtBeginDocument{%
  \providecommand\BibTeX{{%
    \normalfont B\kern-0.5em{\scshape i\kern-0.25em b}\kern-0.8em\TeX}}}

\setcopyright{acmcopyright}
\copyrightyear{2020}
\acmYear{2020}
\acmDOI{10.1145/1122445.1122456}

\acmConference[]{}
\acmBooktitle{}
\acmISBN{}

\usepackage{multirow}

\begin{document}


\title{Is change the only constant? Profile change perspective on \#LokSabhaElections2019}

\author{Kumari Neha}
\affiliation{%
  \institution{IIIT Delhi}
}
\email{nehak@iiitd.ac.in}

\author{Shashank Srikanth}
\affiliation{%
  \institution{IIIT Hyderabad}
}
  \email{shashank.s@research.iiit.ac.in}

\author{Sonali Singhal}
\affiliation{%
 \institution{IIIT Delhi}
}
    \email{sonali18317@iiitd.ac.in}

\author{Shwetanshu Singh}
\affiliation{%
  \institution{IIIT Delhi}
}
\email{shwetanshus@iiitd.ac.in}

\author{Arun Balaji Buduru}
\affiliation{%
  \institution{IIIT Delhi}
}
\email{arunb@iiitd.ac.in}

\author{Ponnurangam Kumaraguru}
\affiliation{%
  \institution{IIIT Delhi}
}
\email{pk@iiitd.ac.in}





\renewcommand{\shortauthors}{Sample Text}
\begin{abstract}
Users on Twitter are identified with the help of their profile attributes that consists of a username, display name, profile image, to name a few. 
The profile attributes that users adopt can reflect their interests, belief, or thematic inclinations. Literature has proposed the implications and significance of profile attribute change for a random population of users. However, the use of profile attribute for endorsements and to start a movement have been under-explored. In this work, we consider \textit{\#LokSabhaElections2019} as a movement and perform a large-scale study of the profile of users who actively made changes to profile attributes centered around \textit{\#LokSabhaElections2019}. We collect the profile metadata for $49.4$M users for a period of $2$ months from April 5, 2019 to June 5, 2019 amid \textit{\#LokSabhaElections2019}. We investigate how the profile changes vary for the influential leaders and their followers over the social movement. We further differentiate the organic and inorganic ways to show the political inclination from the prism of profile changes. We report how the addition of election campaign related keywords lead to spread of behavior contagion and further investigate it with respect to the ``Chowkidar Movement'' in detail. 
\end{abstract}





\maketitle

\section{Introduction}\label{introduction}

User profiles on social media platforms serve as a virtual introduction of the users. People often maintain their online profile space to reflect their likes and values. Further, how users maintain their profile helps them develop relationship with coveted audience~\cite{10.1111/j.1083-6101.2012.01582.x}. A user profile on Twitter is composed of several attributes with some of the most prominent ones being the profile name, screen name, profile image, location, description, followers count, and friend count. While the screen name, display name, profile image, and description identify the user, the follower and friend counts represent the user's social connectivity. 
Profile changes might represent identity choice at a small level. However, previous studies have shown that on a broader level, profile changes may be an indication of a rise in a social movement~\cite{Varol:2014:EOU:2615569.2615699,DBLP:conf/icwsm/WesslenNEGLJS18}. 


In this work, we conduct a large-scale study of the profile change behavior of users on Twitter during the 2019 general elections in India. These elections are of interest from the perspective of social media analysis for many reasons. Firstly, the general elections in India were held in 7 phases from 11th April, 2019 to 19th May, 2019. Thus, the elections serve as a rich source of data for social movements and changing political alignments throughout the two months. Secondly, the increase in the Internet user-base and wider adoption of smartphones has made social media outreach an essential part of the broader election campaign. Thus, there exists a considerable volume of political discourse data that can be mined to gain interesting insights. Twitter was widely used for political discourse even during the 2014 general elections \cite{DBLP:journals/corr/Bhola14} and the number was bound to increase this year. A piece of evidence in favor of this argument is the enormous $49$ million followers that Narendra Modi, the Prime Minister of India has on Twitter\footnote{49M followers as of 27th August, 2019}.
\begin{figure}[!h]
  \centering
  \includegraphics[width=\linewidth]{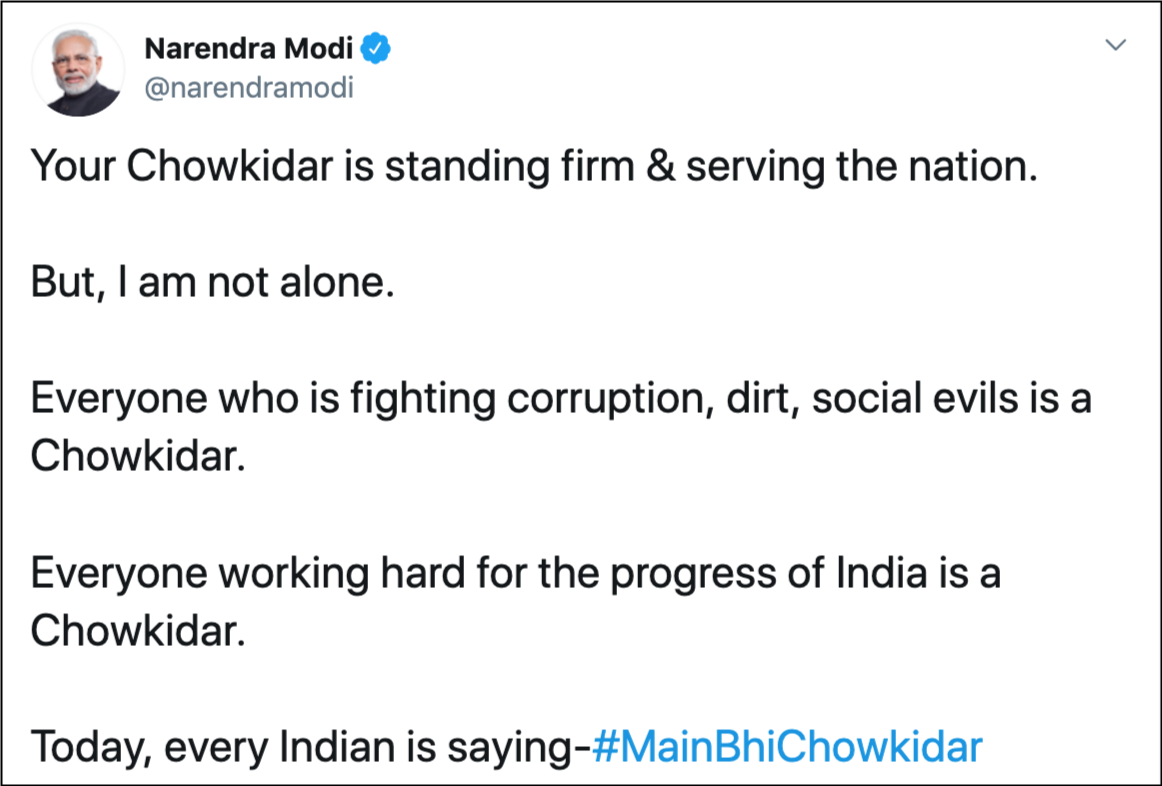}
  \caption{The first occurrence of \#MainBhiChowkidar campaign on 16th March, 2019. One day later, Narendra Modi added \textit{Chowkidar} to his display name on Twitter and kicked off the Chowkidar movement.}
  \label{fig:chowkidar_tweet}
\end{figure}

We believe profile attributes have significant potential to understand how social media played a vital role in the election campaign by political parties as well as the supporters during the \textit{\#LokSabhaElections2019}. A prominent example of the use of profile changes for political campaigning is the \#MainBhiChowkidar campaign launched by the Bhartiya Janta Party (BJP) during the \textit{\#LokSabhaElections2019}. In Figure~\ref{fig:chowkidar_tweet}, we show the first occurrence of \#MainBhiChowkidar campaign on Twitter which was launched in response to an opposition campaign called \textit{``Chowkidar chor hai''} (The watchman is a thief). On 17th March 2019, the Indian Prime Minister Narendra Modi stepped up this campaign and added \textit{Chowkidar} (Watchman) to his display name. This spurred off a social movement on Twitter with users from across the nation and political spectrum updating their profiles on Twitter by prefixing \textit{``Chowkidar''} to their display name \cite{chowkidar}. We are thus interested in studying the different facets of this campaign, in particular, the name changes in the display name of accounts, verified, and others.  

\begin{figure}[!ht]
  \centering
  \includegraphics[width=\linewidth]{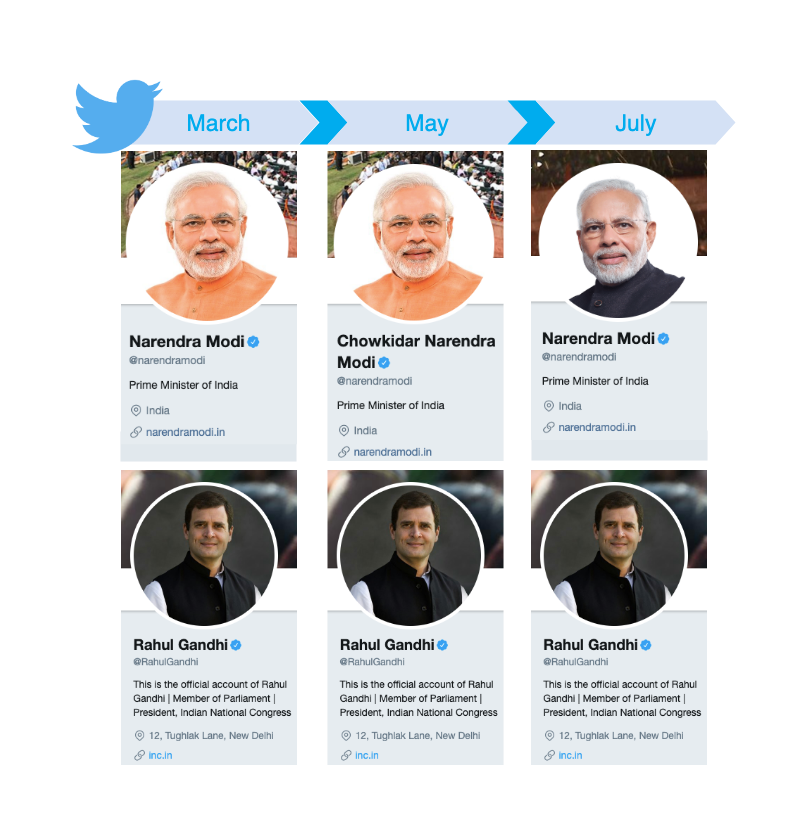}
  \caption{Change in profile attributes of Narendra Modi vs Rahul Gandhi on Twitter. @narendramodi changes his profile attributes frequently as compared to @RahulGandhi on Twitter.}
  \Description{@narendramodi changes his profile attributes frequently as compared to @RahulGandhi on Twitter.}
  \label{fig:chowkidar_sample}
\end{figure}

In Figure~\ref{fig:chowkidar_sample}, we compare the Twitter profiles of the two leaders of national parties during Lok Sabha Election, 2019. The top profile belongs to @narendramodi (Narendra Modi), PM of India and member of the ruling party Bharatiya Janata Party (BJP), while the bottom profile belongs to @RahulGandhi (Rahul Gandhi), president of Indian National Congress (INC). Figure~\ref{fig:chowkidar_sample} shows a snapshot of both leaders before, during, and after the Lok Sabha Elections in 2019. The ruling leader showed brevity in the use of his profile information for political campaigning and changed his profile information during and after the elections, which was not the case with opposition. 


Intending to understand the political discourse from the lens of profile attributes on social media, we address the following questions. 

\subsection{Research Questions}

To analyze the importance of user-profiles in elections, we need to distinguish between the profile change behavior of political accounts and follower accounts. This brings us to the first research question:
\newline
\textbf{RQ1 [Comparison]: } \textit{How often do political accounts and common user accounts change their profile information and which profile attributes are changed the most}. 

The political inclination of Twitter users can be classified based on the content of their past tweets and community detection methods \cite{conover2011predicting}. However, the same analysis has not been extended to the study of user-profiles in detail. We thus, study the following:
\newline
\textbf{RQ2 [Political Inclination]: } \textit{Do users reveal their political inclination on Twitter directly/indirectly using their user profiles only?}

Behavior contagion is a phenomenon where individuals adopt a behavior if one of their opinion leaders adopts it \cite{doi:10.1063/1.5017515}. We analyze the profile change behavior of accounts based on the ``Chowkidar'' movement with the intent to find behavior contagion of an opinion leader. Thus, our third research question is:
\newline
\textbf{RQ3 [Behavior Contagion]:} \textit{Can profile change behavior be catered to behavior contagion effect where followers tend to adopt an opinion as their opinion leader adopts it?} 

We are interested in knowing the common attributes of the people who took part in the \textit{chowkidar} movement. To characterize these users, we need to analyze the major topic patterns that these users engage in. Thus, we ask the question:
\newline
\textbf{RQ4 [Topic Modelling]:} \textit{What are the most discussed topics amongst the users that were part of the Chowkidar campaign?}

\subsection{Contributions}
In summary, our main contributions are:

\begin{itemize}
    \item We collect daily and weekly profile snapshots of $1,293$ political handles on Twitter and over $55$ million of their followers respectively over a period of two months during and after the \#LokSabhaElections2019. We will make the data and code available on our website.
    \item We analyze the political and follower handles for profile changes over the snapshots and show that the political handles engage in more profile change behavior. We also show that certain profile attributes are changed more frequently than others with subtle differences in the behavior of political and follower accounts. 
    \item We show users display support for a political party on Twitter through both organic means like party name mentions on their profile, and inorganic means such as showing support for a movement like \#MainBhiChowkidar. We analyze the \textit{Chowkidar} movement in detail and demonstrate that it illustrates a contagion behavior. 
    \item We perform topic modelling of the users who took part in the \textit{Chowkidar} movement and show that they were most likely Narendra Modi supporters and mostly discussed political topics on Twitter.
\end{itemize}




\section{Dataset Collection and Description}
Our data collection process is divided into two stages. We first manually curate a set of political accounts on Twitter. We use this set of users as a seed set and then collect the profile information of all the followers of these handles.

\textbf{Seed set: } To answer the \textbf{RQ1}, we first manually curate a set of $2,577$ Twitter handles of politicians and official/unofficial party handles involved in Lok Sabha Elections 2019. Let this set of users be called $\mathbf{U}$. Of the $2,577$ curated handles, only $1,293$ were verified at the beginning of April. Let this set of verified handles be called $\mathbf{P}$. We collect a snapshot of the profile information of $\mathbf{P}$ daily for two months from 5th April - 5th June 2019. 

\textbf{Tracked set: } We use the set $P$ of verified Twitter handles as a seed set and collect the profile information of all their followers. We hypothesize that the users who follow one or more political handles are more likely to express their political orientation on Twitter. Thus, we consider only the followers of all the political handles in our work. The total number of followers of all the handles in set $\mathbf{P}$ are $600$ million, of which only $55,830,844$ users are unique. Owing to Twitter API constraints, we collect snapshots of user-profiles for these handles only once a week over the two months of April 5 - June 5, 2019. There were a total of 9 snapshots, of which 7 coincided with the phases of the election. The exact dates on which the snapshots were taken has been mentioned in Table~\ref{table:dataset}. As we collect snapshots over the of two months, the number of followers of political handles in set $P$ can increase significantly. Similarly, a small subset of followers of handles in $P$ could also have been suspended/deleted. To maintain consistency in our analysis, we only use those handles for our analysis that are present in all the $9$ snapshots. We call this set of users as $\mathbf{S}$. The set $\mathbf{S}$ consists of 49,433,640 unique accounts.
 
\textbf{Dataset: } Our dataset consists of daily and weekly snapshots of profile information of users in set $\mathbf{S}$ and $\mathbf{P}$ respectively. We utilize the Twitter API to collect the profile information of users in both the above sets. Twitter API returns a user object that consists of a total of $34$ attributes like name, screen name, description, profile image, location, and followers. 
We further pre-process the dataset to remove all the handles that were inactive for a long time before proceeding with further analysis.

\begin{figure}[!ht]
  \centering
  \includegraphics[width=\linewidth]{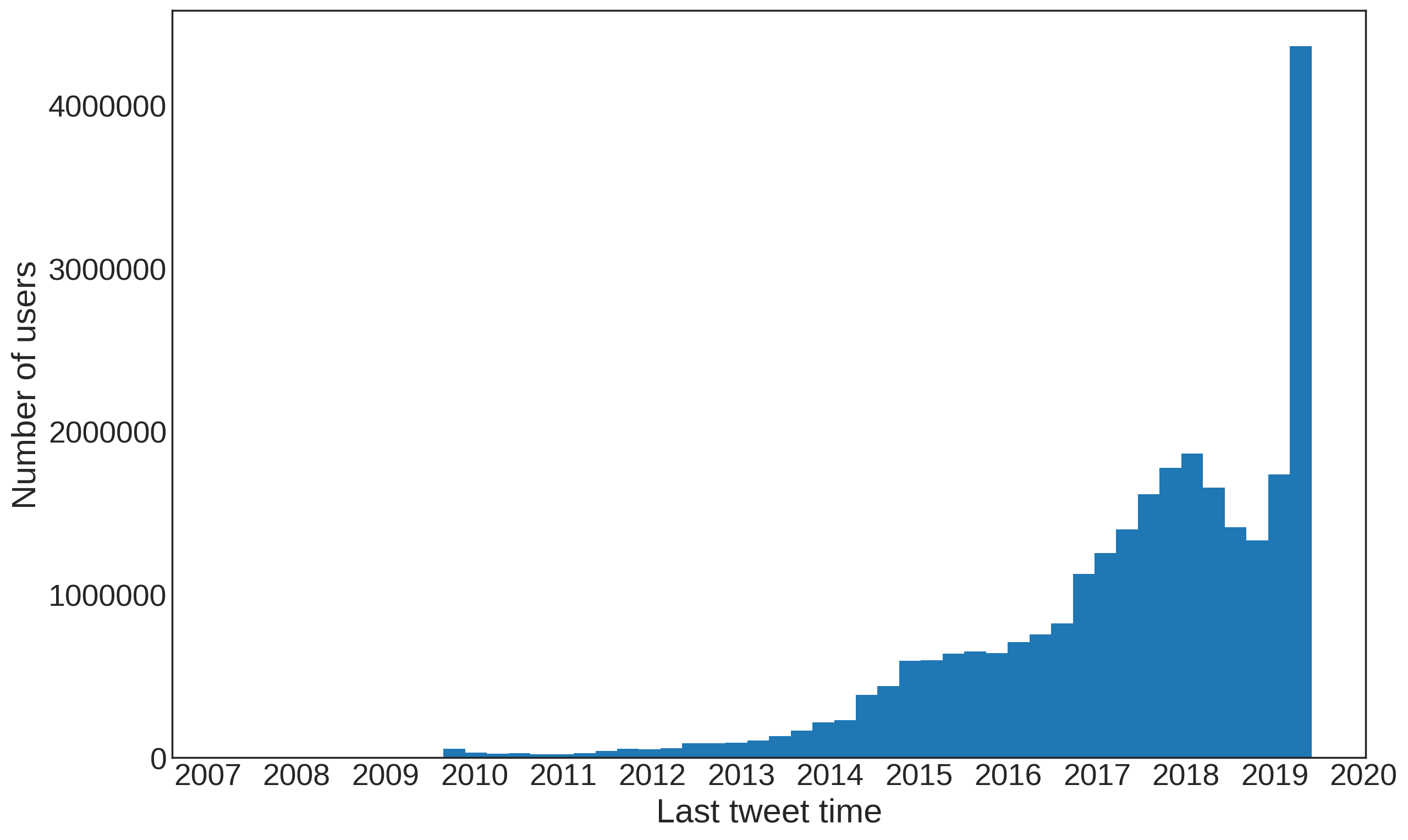}
    \caption{Distribution of last tweet time of twitterer in set $S$. Only $5$ million users tweeted since start of 2019, in favor of hypothesis that a minority of users are only responsible for most of the political opinions.}
  \label{fig:last_tweet}
\end{figure}

In set $\mathbf{S}$, more than $15$ million out of the total $49$ million users have made no tweets and have marked no favourites as well. We call the remaining $34$ million users as Active users. Of the Active users, only $5$ million users have made a tweet in year 2019. The distribution of the last tweet time of all users in set $S$ is shown in Figure~\ref{fig:last_tweet}. The $5$ million twitter users since the start of 2019 fall in favor of the argument that a minority of users are only responsible for most of the political opinions~\cite{Gayo-Avello:2012:NYC:2412375.2412758}. We believe there is a correlation between Active users and users who make changes in profile attribute. We, therefore use the set of $34$ million users(Active Users of set $\mathbf{S}$ ) for our further analysis.

\begin{table}[!ht]
\centering

\begin{tabular}{llc}
\toprule
  & \textbf{Details} &   \\ 
 \midrule
 \multirow{2}{*}{\parbox[c]{1.0cm}{Set $\mathbf{U}$}} 
    & No. of handles & $2,577$ \\
    & No. of political parties & $52$ \\
 \midrule
 \multirow{4}{*}{\parbox[c]{2.3cm}{Set $\mathbf{P}$ (Verified \\  Political Handles)}} 
    & No. of handles & $1,293$ \\
    & Snapshot Frequency & Daily \\
    & No. of snapshots & $61$ \\     
 \midrule
 \multirow{4}{*}{\parbox[c]{2.0cm}{Set $\mathbf{S}$ \\ (Follower Handles)}} 
    & No. of handles & $49,433,640$ \\
    & Snapshot Frequency & Weekly \\    
    & No. of snapshots & $9$ \\
    & Dates of Snapshots & \parbox[c]{2.25cm}{April: $5$, $12$, $20$, $27$ May: $4$, $12$, $20$, $27$ June: $5$} \\
    & No. of verified handles & $11,048$ \\
 \bottomrule
\end{tabular}
\caption{Brief description of the data collection. $U$: seed set of Political handles, $P$: set of verified Political handles, $S$: set of unique Follower handles.}
\label{table:dataset}
\end{table}

\section{Profile Attribute Analysis}

In this section we characterize the profile change behavior of political accounts (Set $P$) and follower accounts (Set $S$). 

\subsection{Political Handles}

Increase in the Internet user-base and widespread adoption of smartphones in the country has led to an increase in the importance of social media campaigns during the elections. This is evident by the fact that in our dataset (Set $U$), we found that a total of $54$ handles got verified throughout $61$ snapshots. Moreover the average number of followers of handles in set $P$ increased from $3,08,716.84$ to $3,19,881.09$ over the $61$ snapshots. Given that the average followers of set $P$ increased by approximately $3\%$ and 54 profile handles got verified in the span of 61 days, we believe that efforts of leaders on social media campaign can't be denied.


We consider $5$ major profile attributes for further analysis. These attributes are \textit{username}, \textit{display name}, \textit{profile image}, \textit{location} and \textit{description} respectively. The \textit{username} is a unique handle or screen name associated with each twitter user that appears on the profile URL and is used to communicate with each other on Twitter. Some username examples are $@narendramodi$, $@RahulGandhi$ etc. The \textit{display name} on the other hand, is merely a personal identifier that is displayed on the profile page of a user, e.g. `Narendra Modi', `Rahul Gandhi', etc. The description is a string to describe the account, while the location is user-defined profile location. We considered the $5$ profile attributes as stated above since these attributes are key elements of a users identity and saliently define users likes and values~\cite{DBLP:conf/icwsm/WesslenNEGLJS18}.

\begin{figure}[!ht]
  \centering
  \includegraphics[width=\linewidth]{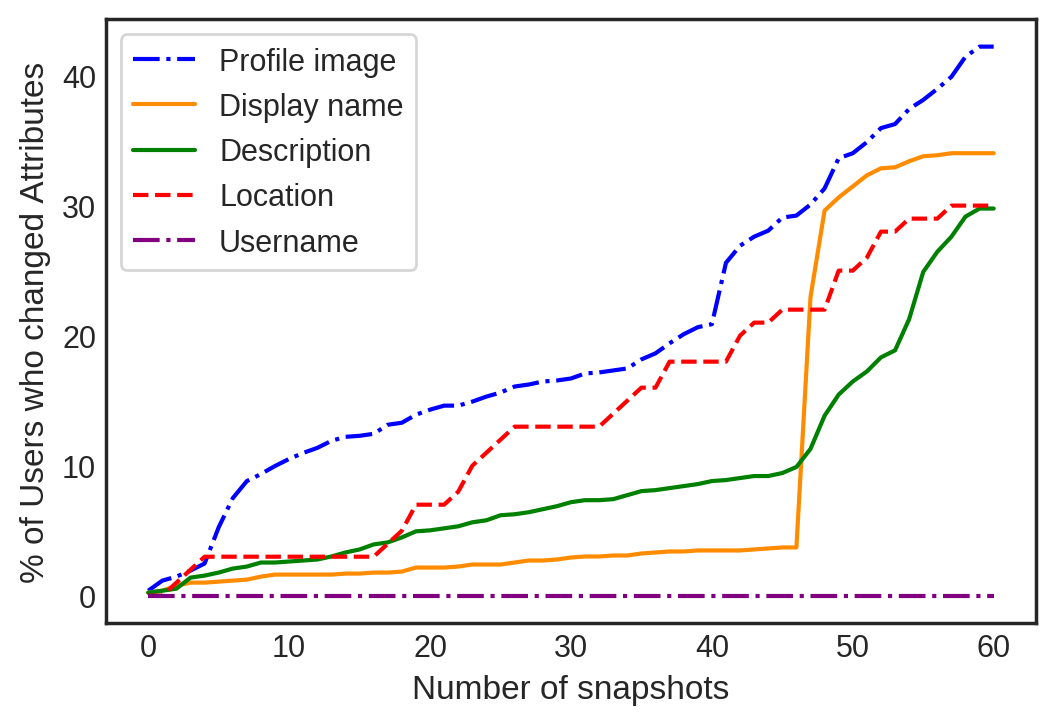}
  \caption{Distribution of profile changes for 5 attributes over 61 snapshots of set $P$. Profile image is the most changed attribute across the snapshots. A sharp increase in \textit{display name} 
coincide with 23rd May, 2019 when \textit{@narendramodi} removed \textit{Chowkidar} from the \textit{display name}.}
  \Description{Distribution of profile changes for 5 attributes over 61 snapshots. Profile image is the most changed attribute across the snapshots. A sharp increase in \textit{display name} 
coincide with May 23, 2019 when \textit{@narendramodi} removed \textit{Chowkidar} from the \textit{display name}.}
  \label{fig:CDF_P_Snapshots}
\end{figure}

For the set of users in $P$, the total number of changes in these $5$ profile attributes over $61$ snapshots is $974$. Figure~\ref{fig:CDF_P_Snapshots} shows the distribution of profile changes for specific attributes over all the 61 snapshots. The most changed attribute among the political handles is the \textit{profile image} followed by  \textit{display name}. In the same Figure, there is a sharp increase in the number of changes to the \textit{display name} attribute on 23rd May, 2019. The increase is because \textit{@narendramodi} removed \textit{Chowkidar} from the \textit{display name} of his Twitter handle which resulted in many other political handles doing the same. Another interesting trend is that none of the handles changed their \textit{username} over all the snapshots. This could be attributed to the fact that Twitter discourages a change of usernames for verified accounts as the verified status might not remain intact after the change~\cite{twitterVerified}. 

\subsection{Follower Handles}

To characterize the profile changing behavior amongst the followers , we study the profile snapshots of all the users in set $S$ over 9 snapshots. 

Of all the active user accounts, $1,363,499$ users made at least one change to their profile attributes over all the snapshots. Thus, over $3\%$ of all the active users engaged in some form of profile change activity over the $9$ snapshots. Of these $1,363,499$ users, more than $95\%$ of the users made only $1$ change to their profile, and around $275$ users made more than $20$ changes. On an average, each one of these $1,363,499$ users made about $2$ changes to their profiles with a standard deviation of $1.42$. 

In order to compare the profile change behavior of political accounts and follower accounts, we analyzed the profile changes of set $\mathbf{P}$ over the $9$ snapshots only. For the same 9 snapshots, $54.9\%$ of all users in set $\mathbf{P}$ made at least one change to their profile attributes. This is in sharp contrast to the the users of set $\mathbf{S}$, where only $3\%$ of users made any changes to their profile attributes. Similarly, users of set $\mathbf{P}$ made $75.32\%$ changes to their profiles on an average, which is 25 times more than the set $\mathbf{S}$.

\begin{figure}[!ht]
  \centering
  \includegraphics[width=\linewidth]{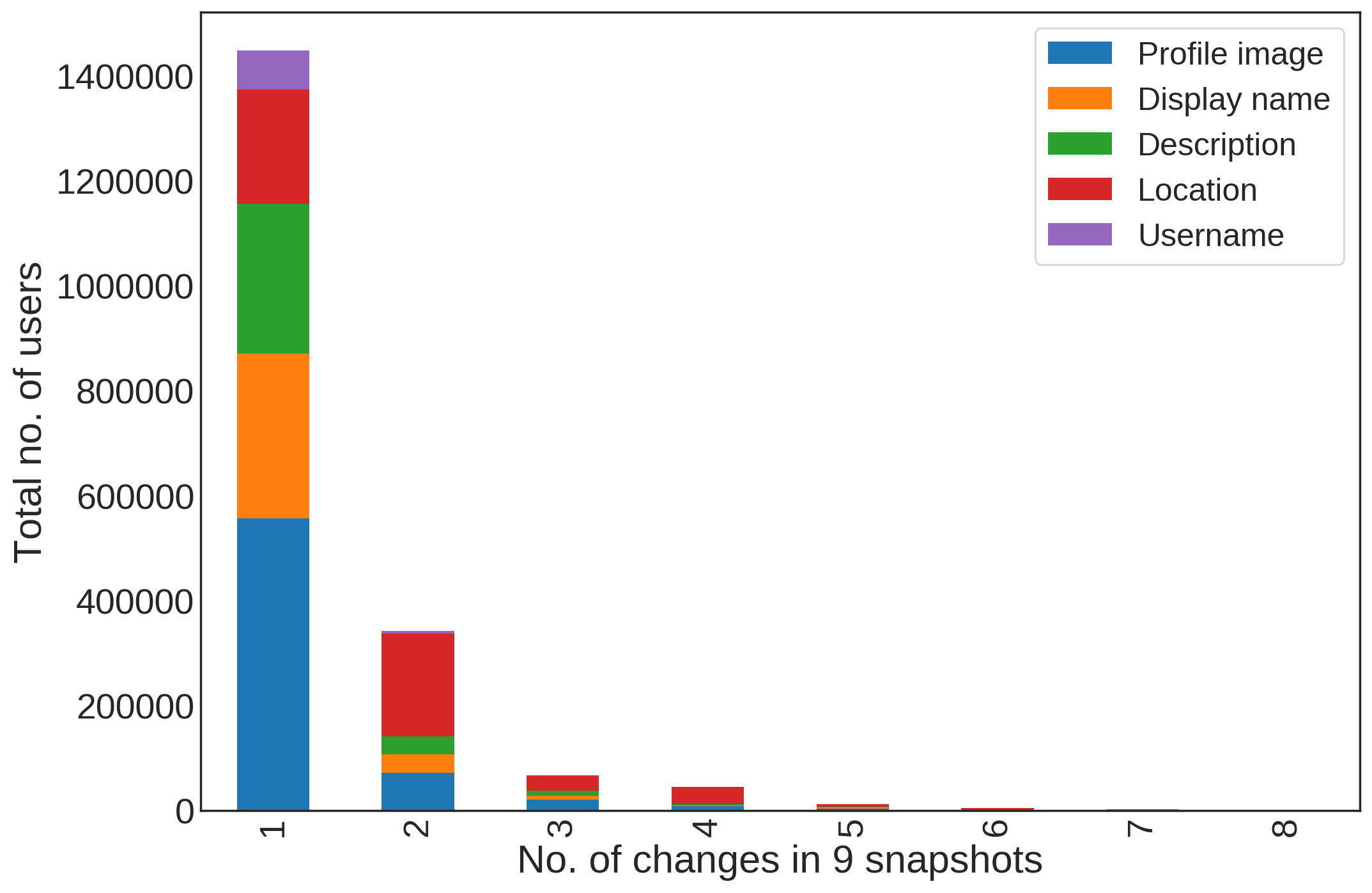}
\caption{Distribution of frequency of change for $5$ attributes over the period of data collection as shown in Table~\ref{table:dataset} for set $S$.}
  \Description{Distribution of frequency of change for $5$ attributes over the period of data collection as shown in Table~\ref{table:dataset} for set $S$.}
  \label{fig:freq_Change}
\end{figure}

\textbf{INSIGHT 1: } \textit{Political handles are more likely to engage in profile changing behavior as compared to their followers.}

In Figure~\ref{fig:freq_Change}, we plot the number of changes in given attributes over all the snapshots for the users in set $S$. From this plot, we find that not all the attributes are modified at an equal rate. \textit{Profile image} and \textit{Location} are the most changed profile attributes and account for nearly $34\%$ and $25\%$ respectively of the total profile changes in our dataset. We analyze the trends in Figure~\ref{fig:CDF_P_Snapshots} and find that the political handles do not change their usernames at all. This is in contrast to the trend in Figure~\ref{fig:freq_Change} where we see that there are a lot of handles that change their usernames multiple times. The most likely reason for the same is that most of the follower handles are not verified and would not loose their verified status on changing their \textit{username}.

The follower handles also use their \textit{usernames} to show support to a movement or person and revert back to their original \textit{username} later. Out of the $79,991$ \textit{username} changes, $1,711$ users of set $\boldsymbol{S}$ went back and forth to the same name which included adding of Election related keywords. Examples of few such cases in our dataset are shown in Table~\ref{tab:username_change}. 

\begin{table}[ht]
\centering
\begin{tabular}{lll} 
  \hline
  \textbf{Username at $T_0$} & \textbf{Username at $T_1$} & \textbf{Username at $T_2$} \\
  \hline
  prahld\_kushwaha & ChowkidarPrahld & prahld\_kushwaha \\ 
  Nammamodi1 & DcpShivaraj19 & Nammamodi1 \\
 \hline
\end{tabular}
\caption{Example of users who changed their username once and later reverted back to their old username.}
\label{tab:username_change}
\end{table}

\textbf{INSIGHT 2: } \textit{Users do not change all profile attributes equally, and the profile attributes that political and follower handles focus on are different.}

\subsection{Similarity analysis}
Change in an attribute in the profile involves replacing an existing attribute with a new one. We analyze the similarity between the current, new \textit{display names} and \textit{usernames} used by users in set $\mathbf{P}$ and $\mathbf{S}$ respectively. For this analysis, we consider only the users who have changed their \textit{username} or \textit{display name} at least once in all $9$ snapshots. We use the Longest Common Subsequence (LCS) to compute the similarity between two strings. Thus, for each user, we compute the average LCS score between all possible pairs of unique \textit{display names} and \textit{usernames} adopted by them. A high average LCS score indicates that the newer \textit{username} or \textit{display name} were not significantly different from the previous ones. 

\begin{figure}[!ht]
  \centering
  \includegraphics[width=\linewidth]{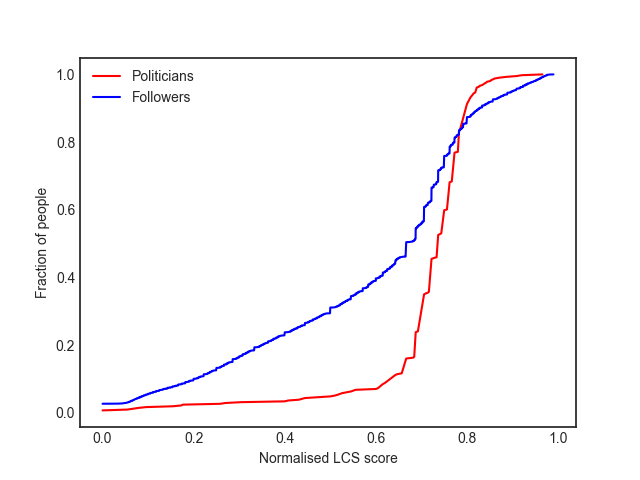}
  \caption{Distribution of $Display$ $name$ change among set $S$ and $P$. Higher Longest Comman Subsequence(LCS) score indicates the old and new $Display$ $names$ were similar or related.}
  \Description{Analysis of Display Name changes for Set P and Set S}
  \label{fig:LCS_cdf}
\end{figure}

The graph in Figure~\ref{fig:LCS_cdf} shows that nearly $80\%$ of the users in set $S$ made more significant changes to their Display Names as compared to the political handles. (This can be seen by considering a horizontal line on the graph. For any given fraction of users, the average LCS score of set $S$ (Followers) is lesser than set $P$ (Politicians) for approximately $80\%$ percent of users). For around $10\%$ of the users in set $P$, the new and old profile attributes were unrelated (LCS score < $0.5$) and for users in set $S$, this value went up to $30\%$. 

\textbf{INSIGHT 3:} \textit{Political handles tend to make new changes related to previous attribute values. However, the followers make comparatively less related changes to previous attribute values.}

\section{Political support through profile meta data}



In this section, we first describe the phenomenon of mention of political parties names in the profile attributes of users. This is followed by the analysis of profiles that make specific mentions of political handles in their profile attributes. Both of these constitute an organic way of showing support to a party and does not involve any direct campaigning by the parties. We also study in detail the \#MainBhiChowkidar campaign and analyze the corresponding change in profile attributes associated with it. 

\subsection{Party mention in the profile attributes}\label{party_mention_section}

To sustain our hypothesis of party mention being an organic behavior, we analyzed party name mentions in the profiles of users in set $\mathbf{S}$. For this analysis, we focused on two of the major national parties in India, namely Indian National Congress (INC) and Bhartiya Janta Party (BJP). We specifically search for mentions of the party names and a few of its variants in the \textit{display name}, \textit{username} and \textit{description} of the user profiles. We find that $50,432$ users have mentioned ``BJP'' (or its variants) on their \textit{description} in the first snapshot of our data itself. This constitutes approximately $1\%$ of the total number of active users in our dataset, implying a substantial presence of the party on Twitter. Moreover, $2,638$ users added ``BJP'' (or its variants) to their \textit{description} over the course of data collection and $1,638$ users removed ``BJP'' from their \textit{description}. Table~\ref{table:bjpIncPartyNames} shows the number of mentions of both parties in the different profile attributes. 


\begin{table}[!ht]
\centering
\begin{tabular}{llllc}
\toprule
 \parbox[c]{1cm}{\textbf{Profile \\ Attribute}} & \parbox[c]{1cm}{\textbf{Party \\ Name}} & \textbf{Present} & \textbf{Addition} & \textbf{Removal}  \\ 
 \midrule
 \multirow{2}{*}{\textbf{Description}}
    & BJP & $50,431$ & $2,638$ & $1,638$ \\
    & INC & $8,967$ & $408$ & $325$ \\
 \midrule
 \multirow{2}{*}{\textbf{Display Name}} 
    & BJP & $15,305$ & $996$ & $506$ \\
    & INC & $8,692$ & $204$ & $179$ \\ 
 \bottomrule
 \multirow{2}{*}{\textbf{Username}} 
    & BJP & $17855$ & $314$ & $163$ \\
    & INC & $7042$ & $104$ & $58$ \\ 
 \bottomrule 
\end{tabular}
\caption{List for frequency of presence, addition and removal of party names to given profile attributes in the dataset.}
\label{table:bjpIncPartyNames}
\end{table}

\subsection{Who follows who}

We observe that a lot of users in our dataset wrote that they are proud to be followed by a famous political handle in their description. We show example of such cases in Table~\ref{tab:username_change_description}.

 We perform an exhaustive search of all such mentions in the description attribute of the users and find $1,164$ instances of this phenomenon. This analysis and the party mention analysis in Section~\ref{party_mention_section}, both testify that people often display their political inclination on Twitter via their profile attributes itself.

\textbf{INSIGHT 4: } \textit{Twitter users often display their political inclinations in their profile attributes itself and are pretty open about it.}

\begin{table}[ht]
\centering
\begin{tabular}{p{2cm}p{5cm}}
  \hline
  \textbf{Username} & \textbf{Description} \\
  \hline
   ravibhadoria & Honoured to be followed by PM @narendramodi ji and worlds largest party's president @amitshah ji. RTs Not Endorsement... My Tweets In Likes... \\ 
   \midrule
   Sushant\_Kaushal & Businessman | Ardent Supporter of Narendra Modi | Swyamsevak| Socio-Political Analyst| Outspoken | TeamModiOnceMore| Blessed to Be Followed By @NarendraModi ji \\
 \hline
\end{tabular}
\caption{Example of users who wrote followed by a famous politician in their description}
\label{tab:username_change_description}
\end{table}

\subsection{The Chowkidar movement}

\begin{figure}[ht]
  \centering
  \includegraphics[width=\linewidth]{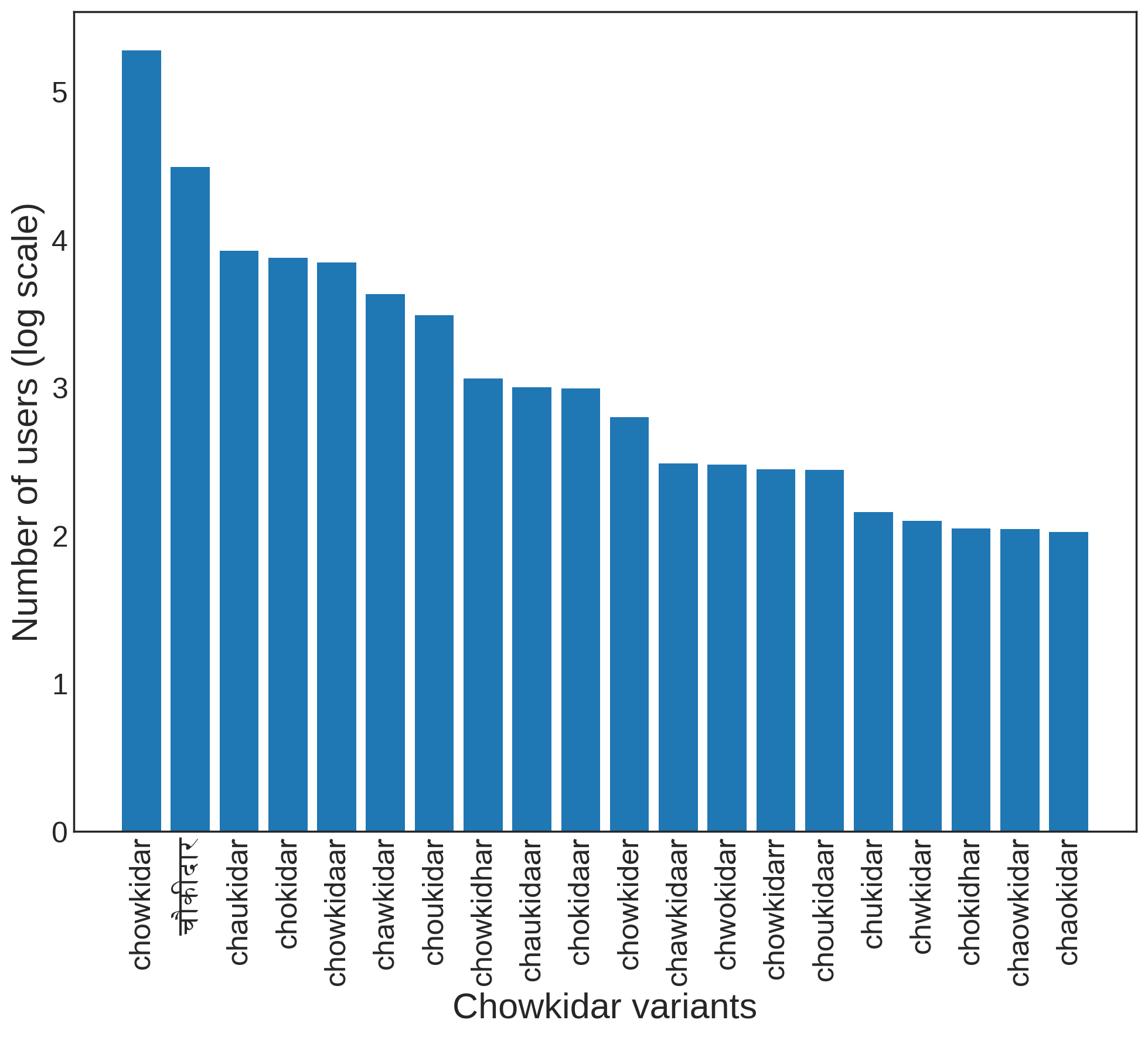}
  \caption{Most popular variants of Chowkidar and their corresponding frequency of use.}
  \Description{Several variants of Chowkidar and their corresponding frequency of use.}
  \label{fig:chowkidar_variants}
\end{figure}

As discussed is Section~\ref{introduction}, @narendramodi added ``\textit{Chowkidar}'' to his display name in response to an opposition campaign called \textit{``Chowkidar chor hai''}. In a coordinated campaign, several other leaders and ministers of BJP also changed their Twitter profiles by adding the prefix \textit{Chowkidar} to their display names. The movement, however, did not remain confined amongst the members of the party themselves and soon, several Twitter users updated their profiles as well. 
We opine that the whole campaign of adding \textit{Chowkidar} to the profile attributes show an inorganic behavior, with political leaders acting as the catalyst.
An interesting aspect of this campaign was the fact that the users used several different variants of the word \textit{Chowkidar} while adding it to their Twitter profiles. Some of the most common variants of \textit{Chowkidar} that were present in our dataset along with its frequency of use is shown in Figure~\ref{fig:chowkidar_variants}. 

We utilize a regular expression query to exhaustively search all the variants of the word \textit{Chowkidar} in our dataset. We found $2,60,607$ instances of users in set $S$ added \textit{Chowkidar} (or its variants) to their display names. These $2,60,607$ instances comprised a total of $241$ unique chowkidar variants of which $225$ have been used lesser than $500$ times. We also perform the same analysis for the other profile attributes like description and username as well. The number of users who added \textit{Chowkidar} to their description and username are $14,409$ and $12,651$ respectively. The union and intersection of all these users are $270,945$ and $727$ respectively, implying that most users added \textit{Chowkidar} to only one of their profile attributes.

We believe, the effect of changing the profile attribute in accordance with Prime Minister's campaign is an example of inorganic behavior contagion \cite{doi:10.1063/1.5017515,doi:10.1002/ejsp.2615}. The authors in \cite{doi:10.1063/1.5017515} argue that opinion diffuses easily in a network if it comes from opinion leaders who are considered to be users with a very high number of followers. We see a similar behavior contagion in our dataset with respect to the \textit{Chowkidar} movement. 

We analyze the similarity of the display names with respect to the behavior contagion of \textit{Chowkidar} movement. In the CDF plot of Figure~\ref{fig:LCS_cdf}, a significant spike is observed in the region of LCS values between $0.6$-$0.8$. This spike is caused mostly due to the political handles, who added \textit{Chowkidar} to their display names which accounted for $95.7\%$ of the users in this region. In set $P$, a total of $373$ people added the specific keyword \textit{Chowkidar} to their display names of which 315 lie in the normalized LCS range of $0.6$ to $0.8$. We perform similar analysis on the users of set $S$ and find that around $57\%$ of the users within the region of $0.6$-$0.8$ LCS added specific keyword \textit{Chowkidar} to their display names. We perform similar analysis on other campaign keywords like \textit{Ab Hoga Nyay} as well but none of them have the same level of influence or popularity like that of the ``Chowkidar'' and hence we call it as the ``Chowkidar Movement''.

\begin{figure}[ht] 
  \centering
  \fbox{\includegraphics[width=0.9\linewidth]{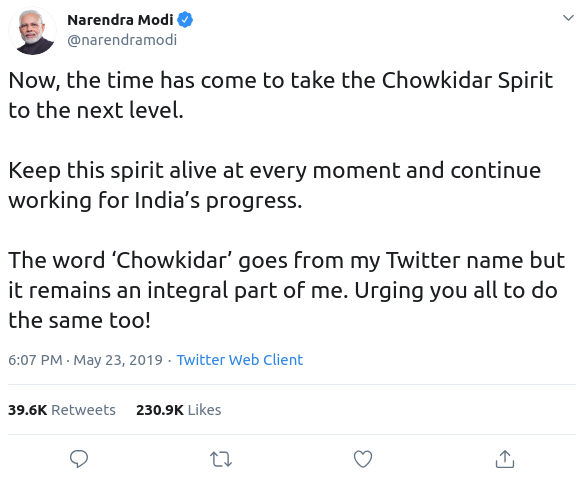}}
  \caption{The tweet in which Narendra Modi urges Twitter users to remove \textit{Chowkidar} from their profiles.}
  \Description{The tweet in which Narendra Modi urges Twitter users to remove \textit{Chowkidar} from their profiles.}
  \label{fig:chowkidar_remove}
\end{figure}

\begin{figure}[ht] 
  \centering
  \includegraphics[width=\linewidth]{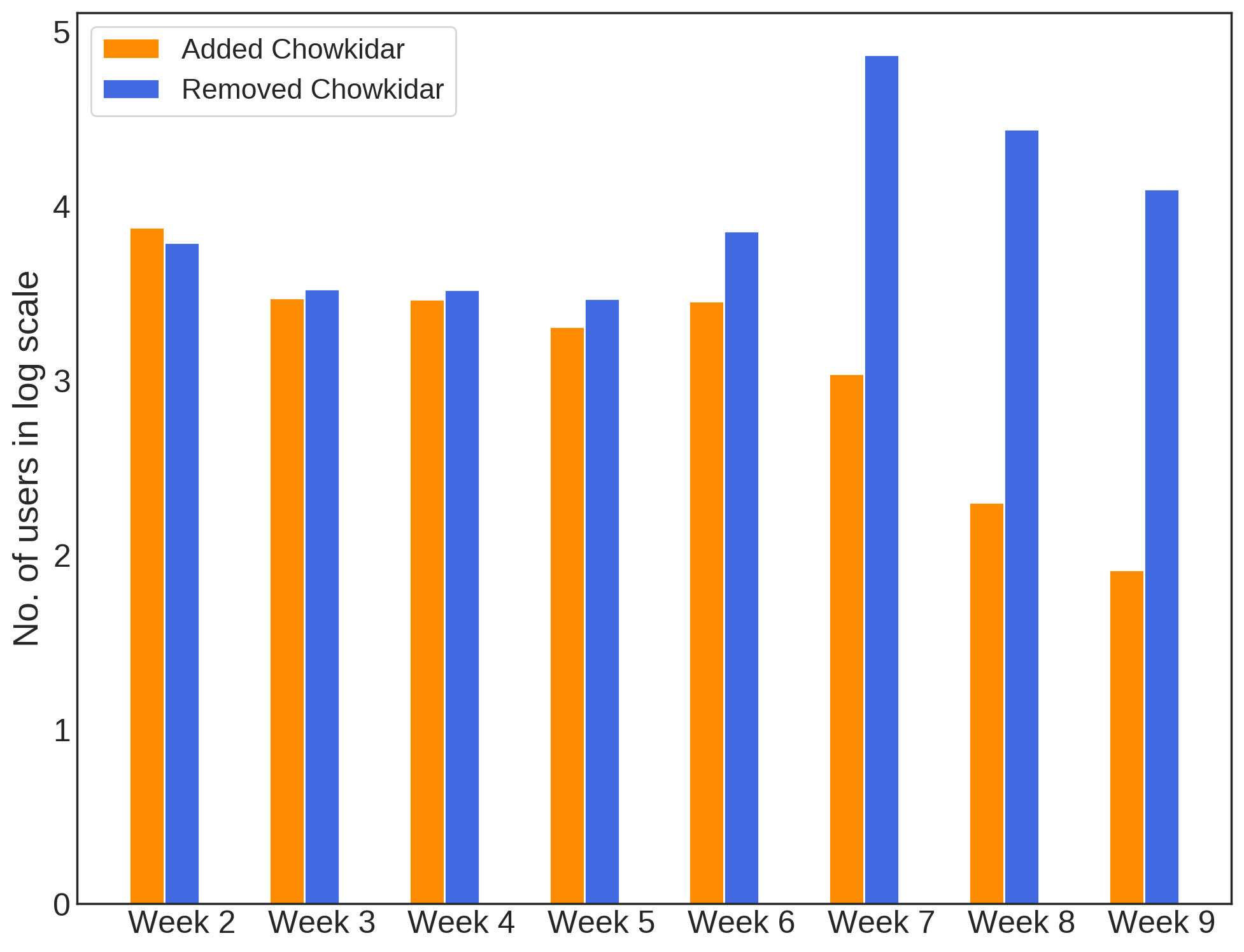}
  \caption{Weekly addition and deletion of Chowkidar to profile attributes.}
  \Description{Weekly addition and deletion of Chowkidar to profile attributes.}
  \label{fig:chowkidar_addition}
\end{figure}

On $23$rd May, Narendra Modi removed \textit{Chowkidar} from his Twitter profile and urged others to do the same as shown in Figure~\ref{fig:chowkidar_remove}. While most users followed Narendra Modi's instructions and removed \textit{Chowkidar} from their profiles, some users still continued to add \textit{chowkidar} to their names. The weekly addition and removal of \textit{Chowkidar} from user profiles is presented in detail in Figure~\ref{fig:chowkidar_addition}. Thus, the behavior contagion is evident from the fact that after Narendra Modi removed \textit{Chowkidar} on 23rd May, majority of the population in set $\mathbf{S}$ removed it too. 

\textbf{INSIGHT 5: } The \textit{Chowkidar} movement was very widespread and illustrated contagion behavior.

\section{Topic modelling of users}

\begin{figure}[!ht] 
  \centering
  \includegraphics[width=\linewidth]{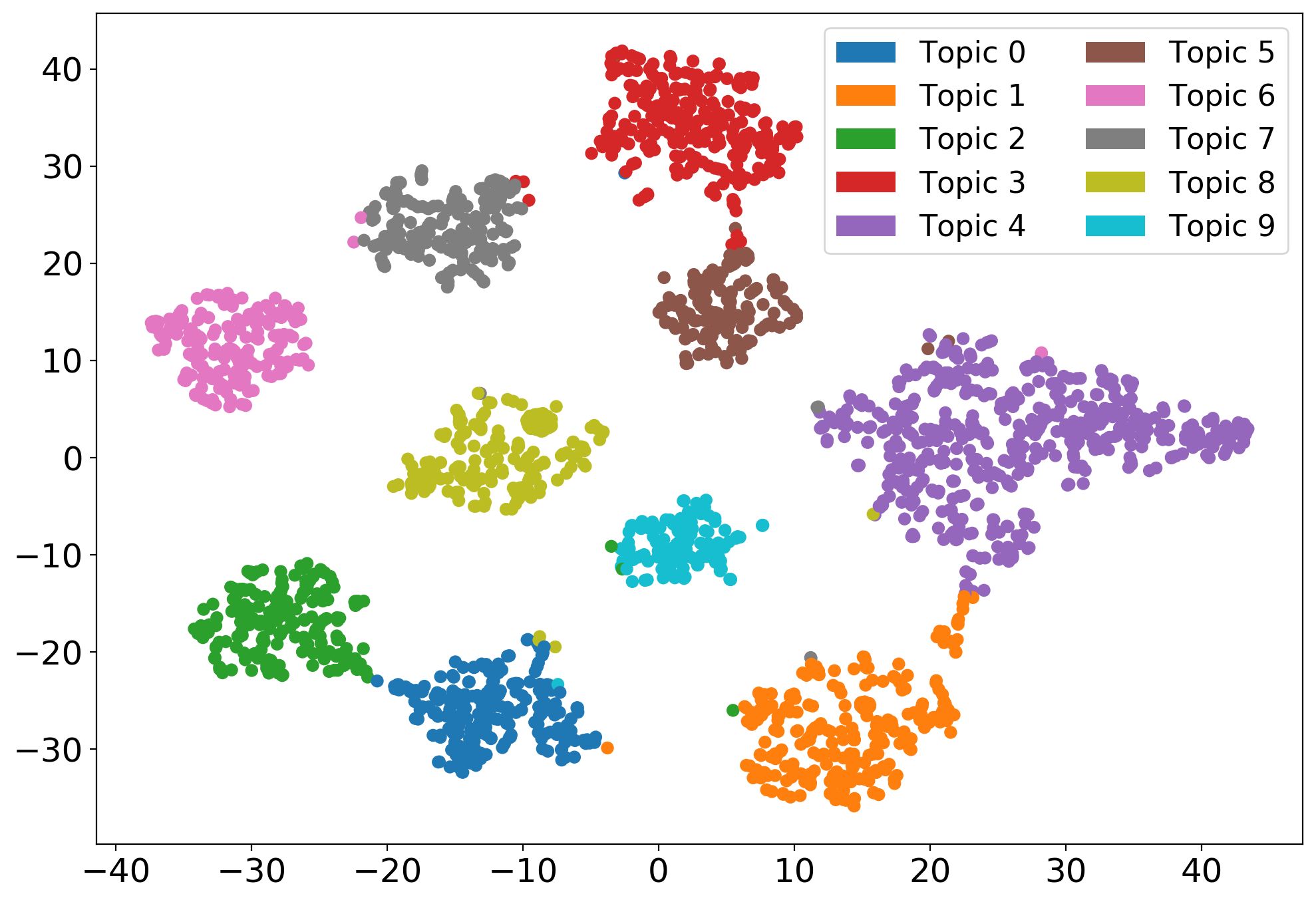}
  \caption{Clusters of the topics on the 2D plane after performing tSNE on the data.}
  \Description{Clusters of the topics in 2D plane after performing tSNE on the data.}
  \label{fig:topic_clusters}
\end{figure}

A large number of people were part of the \textit{Chowkidar} movement. Hence, it is important to gauge their level of interest/support to the party in order to accurately judge the impact of the campaign. We thus perform topic modelling of the description attribute of these users. We utilize the LDA Mallet model~\cite{McCallumMALLET} for topic modelling after performing the relevant preprocessing steps on the data. We set the number of topics to $10$ as it gave the highest coherence score in our experiments. The $10$ topics and the $5$ most important keywords belonging to these topics are shown below in Figure~\ref{fig:topic_modelling}. These topics reveal several interesting insights. The most discussed topic amongst these users involves Narendra Modi and includes words like ``namo'' (slang for Narendra Modi), supporter, nationalist and ``bhakt'' (devout believer). A few other topics also involve politics like topic $2$ and topic $5$. Topic $2$, in particular contains the words \textit{chowkidar}, student, and study, indicating that a significant set of users who were part of this campaign were actually students. Topic number $0$ accounts for more than $40\%$ of the total descriptions, which essentially means that most of the people involved in this movement were supporting BJP or Narendra Modi. We also plot the topics associated with each user's description on a 2D plane by performing tSNE~\cite{sklearn_api} on the $10$ dimensional data to ascertain the variation in the type of topics. The topics are pretty distinct and well separated on the $2$D plane as can be seen in Figure~\ref{fig:topic_clusters}. 

\begin{figure}[ht] 
  \centering
  \includegraphics[width=\linewidth]{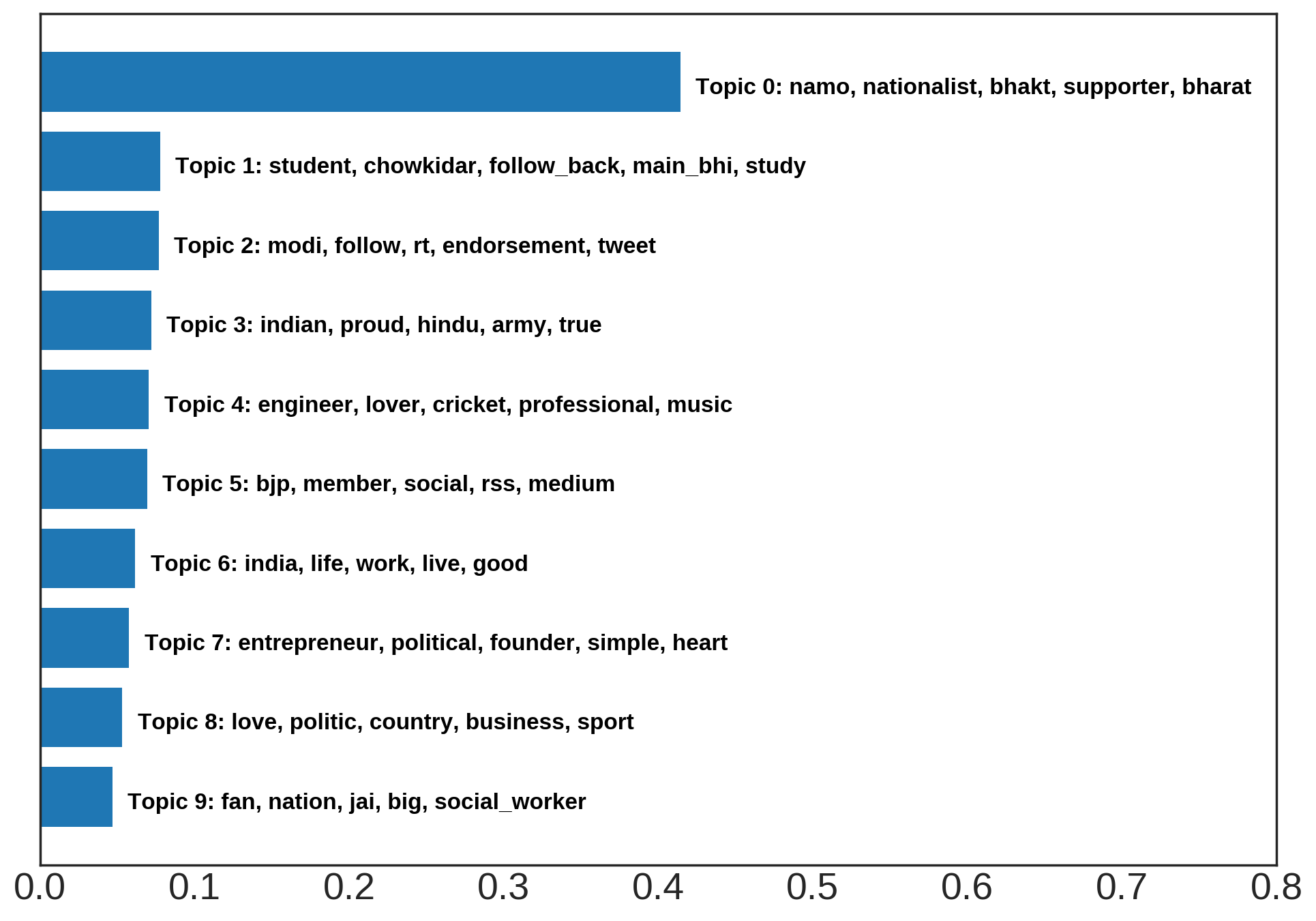}
  \caption{The most discussed topics in a descending order along with the $5$ most representative words of these topics.}
  \Description{The most discussed topics in a descending order along with the $5$ most representative words of these topics.}
  \label{fig:topic_modelling}
\end{figure}

\textbf{INSIGHT 6: } \textit{The users who were part of the chowkidar movement were mostly Narendra Modi supporters and engaged in political topics}

\section{Related Work} \label{relatedWwork}
Social Media profiles are a way to present oneself and to cultivate relationship with like minded or desired audience.~\cite{10.1111/j.1083-6101.2012.01582.x,TheEffectofSocialMediaonIdentityConstruction}.
With change in time and interest, users on Twitter change their profile names, description, screen names, location or language~\cite{DBLP:conf/icwsm/WesslenNEGLJS18}. Some of these changes are organic~\cite{Mariconti:2017:WNU:3038912.3052589,DBLP:conf/icwsm/WesslenNEGLJS18}, while other changes can be triggered by certain events~\cite{Khatua:2019:TSL:3297001.3297057,doi:10.1080/1369118X.2017.1301522}. In this section, we look into the previous literature which are related to user profile attribute's change behavior. We further look into how Twitter has been used to initiate or sustain movements~\cite{Khatua:2019:TSL:3297001.3297057,doi:10.1080/1369118X.2017.1301522,Varol:2014:EOU:2615569.2615699}. Finally we investigate how the contagion behavior adaptation spread through opinion leaders~\cite{doi:10.1063/1.5017515,doi:10.1002/ejsp.2615} and conclude with our work's contribution to the literature.

\subsection{Profile attribute change behavior}
Previous studies have focused on change in username attribute extensively, as usernames in Twitter helps in linkability, mentions and becomes part of the profile's URL~\cite{Jain:2016:DUC:2888451.2888452,Mariconti:2017:WNU:3038912.3052589,Mariconti:2016:WAP:2905760.2905762}. While the focus of few works is to understand the reason the users change their profile names~\cite{Jain:2016:DUC:2888451.2888452,DBLP:conf/icwsm/WesslenNEGLJS18}, others strongly argue username change is a bad idea~\cite{Mariconti:2016:WAP:2905760.2905762,Mariconti:2017:WNU:3038912.3052589}. Jain et al. argue that the main reason people opt to change their username is to leverage more tweet space. Similarly, the work by Mariconti et al. suggests that the change in username may be a result of name squatting, where a person may take away the username of a popular handle for malicious reasons. 
More recent studies in the area have focused on several profile attributes changes ~\cite{DBLP:conf/icwsm/WesslenNEGLJS18}. They suggest that profile attributes can be used to represent oneself at micro-level as well as represent social movement at macro-level. 


\subsection{Social Media for online movements}
The role of Social Media during social movements have been studied very extensively~\cite{Khatua:2019:TSL:3297001.3297057,morstatter2018alt,johansson2018opening,doi:10.1080/1369118X.2017.1301522}. 
With the help of Social Media, participation and ease to express oneself becomes easy~\cite{Varol:2014:EOU:2615569.2615699}. 
There have been many studies where Twitter has been used to gather the insight of online users in the form of tweets~\cite{Varol:2014:EOU:2615569.2615699,Khatua:2019:TSL:3297001.3297057,doi:10.1080/1369118X.2017.1301522}. The authors in ~\cite{Khatua:2019:TSL:3297001.3297057} study the tweets to understand how people reacted to decriminalization of LGBT in India. The study of Gezi Park protest shows how the protest took a political turn, as people changed their usernames to show support to their political leaders~\cite{Varol:2014:EOU:2615569.2615699}. 


The success of the party ``Alternative for Deutschland(AFD)'' in German Federal Elections, 2017 can be catered to its large online presence~\cite{morstatter2018alt}. The authors in \cite{morstatter2018alt} study the tweets and retweets networks during the German elections and analyze how users are organized into communities and how information flows between them. The role of influential leaders during elections has also been studied in detail in studies like ~\cite{bovet2019influence}. This brings us to the study of how influential leaders contribute to spread and adaptation of events.



\subsection{Adoption of behavior from leaders}
The real and virtual world are intertwined in a way that any speech, event, or action of leaders may trigger the changes is behavior of people online~\cite{Varol:2014:EOU:2615569.2615699,PARK20131641}. 
Opinion leaders have been defined as `the individuals who were likely to influence other persons in their immediate environment.'~\cite{katzlazarsfeld}
Opinion leaders on social media platforms help in adoption of behavior by followers  easily~\cite{doi:10.1002/ejsp.2615,doi:10.1063/1.5017515}. The performance and effectiveness of the opinion leader directly affect the contagion of the leader~\cite{doi:10.1002/ejsp.2615}. On Twitter, opinion leaders show high motivation for information seeking and public expression. Moreover, they make a significant contribution to people's political involvement~\cite{PARK20131641}. 


Our work contributes to this literature in two major ways. First, we investigate how profile attribute changes are similar or different from the influential leaders in the country and how much of these changes are the result of the same. Then we see how the users change their profile information in the midst of election in the largest democracy in the world.

\section{Discussion}
In this paper, we study the use of profile attributes on Twitter as the major endorsement during \textit{\#LokSabhaElectios2019}. We identify $1,293$ verified political leader's profile and extract $49,433,640$ unique  followers. We collect the daily and weekly snapshots of the political leaders and the follower handles from 5th April, 2019 to 5th June, 2019. With the constraint of Twitter API, we collected a total of 9 snapshots of the follower's profile and 61 snapshots of Political handles. We consider the 5 major attributes for the analysis, including, \textit{Profile Name, Username, Description, Profile Image}, and \textit{Location}. 

Firstly, we analyze the most changed attributes among the 5 major attributes. We find that most of the users made at-most $4$ changes during the election campaign, with profile image being the most changed attribute for both the political as well as follower handles. We also show that political handles made more changes to profile attributes than the follower handles. About $80\%$ changes made by the followers in the period of data collection was related to \textit{\#LokSabhaElectios2019}. While the profile handles made changes, the changes mostly included going back and forth to same profile attribute, with the change in attribute being active support to a political party or political leader.

We argue that we can predict the political inclination of a user using just the profile attribute of the users. We further show that the presence of party name in the profile attribute can be considered as an organic behavior and signals support to a party. However, we argue that the addition of election campaign related keywords to the profile is a form of inorganic behavior. The inorganic behavior analysis falls inline with behavior contagion, where the followers tend to adapt to the behavior of their opinion leaders. The ``Chowkidar movement'' showed a similar effect in the \textit{\#LokSabhaElectios2019}, which was evident by how the other political leaders and followers of BJP party added chowkidar to their profile attributes after @narendramodi did.  We thus, argue that people don't shy away from showing support to political parties through profile information. 

We believe our analysis will be helpful to understand how social movement like election are perceived and sustained through profile attributes on social media. Our findings provide baseline data for study in the direction of election campaign and to further analyze people's perspective on elections.
\bibliographystyle{ACM-Reference-Format}
\bibliography{sample-base}


\begin{thebibliography}{23}


\ifx \showCODEN    \undefined \def \showCODEN     #1{\unskip}     \fi
\ifx \showDOI      \undefined \def \showDOI       #1{#1}\fi
\ifx \showISBNx    \undefined \def \showISBNx     #1{\unskip}     \fi
\ifx \showISBNxiii \undefined \def \showISBNxiii  #1{\unskip}     \fi
\ifx \showISSN     \undefined \def \showISSN      #1{\unskip}     \fi
\ifx \showLCCN     \undefined \def \showLCCN      #1{\unskip}     \fi
\ifx \shownote     \undefined \def \shownote      #1{#1}          \fi
\ifx \showarticletitle \undefined \def \showarticletitle #1{#1}   \fi
\ifx \showURL      \undefined \def \showURL       {\relax}        \fi
\providecommand\bibfield[2]{#2}
\providecommand\bibinfo[2]{#2}
\providecommand\natexlab[1]{#1}
\providecommand\showeprint[2][]{arXiv:#2}

\bibitem[\protect\citeauthoryear{Bhola}{Bhola}{2014}]%
        {DBLP:journals/corr/Bhola14}
\bibfield{author}{\bibinfo{person}{Abhishek Bhola}.}
  \bibinfo{year}{2014}\natexlab{}.
\newblock \bibinfo{title}{Twitter and Polls: Analyzing and estimating political
  orientation of Twitter users in India General {\#}Elections2014}.
\newblock
\newblock
\showeprint[arxiv]{1406.5059}
\urldef\tempurl%
\url{http://arxiv.org/abs/1406.5059}
\showURL{%
\tempurl}


\bibitem[\protect\citeauthoryear{Bovet and Makse}{Bovet and Makse}{2019}]%
        {bovet2019influence}
\bibfield{author}{\bibinfo{person}{Alexandre Bovet} {and}
  \bibinfo{person}{Hern{\'a}n~A Makse}.} \bibinfo{year}{2019}\natexlab{}.
\newblock \showarticletitle{Influence of fake news in Twitter during the 2016
  US presidential election}.
\newblock \bibinfo{journal}{\emph{Nature communications}} \bibinfo{volume}{10},
  \bibinfo{number}{1} (\bibinfo{year}{2019}), \bibinfo{pages}{7}.
\newblock


\bibitem[\protect\citeauthoryear{Buitinck, Louppe, Blondel, Pedregosa, Mueller,
  Grisel, Niculae, Prettenhofer, Gramfort, Grobler, Layton, VanderPlas, Joly,
  Holt, and Varoquaux}{Buitinck et~al\mbox{.}}{2013}]%
        {sklearn_api}
\bibfield{author}{\bibinfo{person}{Lars Buitinck}, \bibinfo{person}{Gilles
  Louppe}, \bibinfo{person}{Mathieu Blondel}, \bibinfo{person}{Fabian
  Pedregosa}, \bibinfo{person}{Andreas Mueller}, \bibinfo{person}{Olivier
  Grisel}, \bibinfo{person}{Vlad Niculae}, \bibinfo{person}{Peter
  Prettenhofer}, \bibinfo{person}{Alexandre Gramfort}, \bibinfo{person}{Jaques
  Grobler}, \bibinfo{person}{Robert Layton}, \bibinfo{person}{Jake VanderPlas},
  \bibinfo{person}{Arnaud Joly}, \bibinfo{person}{Brian Holt}, {and}
  \bibinfo{person}{Ga{\"{e}}l Varoquaux}.} \bibinfo{year}{2013}\natexlab{}.
\newblock \showarticletitle{{API} design for machine learning software:
  experiences from the scikit-learn project}. In \bibinfo{booktitle}{\emph{ECML
  PKDD Workshop: Languages for Data Mining and Machine Learning}}.
  \bibinfo{pages}{108--122}.
\newblock


\bibitem[\protect\citeauthoryear{Clarkson, Wagstaff, Arthur, and
  Thelwell}{Clarkson et~al\mbox{.}}{2019}]%
        {doi:10.1002/ejsp.2615}
\bibfield{author}{\bibinfo{person}{Beth~G. Clarkson},
  \bibinfo{person}{Christopher~R.D. Wagstaff}, \bibinfo{person}{Calum~A.
  Arthur}, {and} \bibinfo{person}{Richard~C. Thelwell}.}
  \bibinfo{year}{2019}\natexlab{}.
\newblock \showarticletitle{Leadership and the contagion of affective
  phenomena: A systematic review and mini meta-analysis}.
\newblock \bibinfo{journal}{\emph{European Journal of Social Psychology}}
  \bibinfo{volume}{0}, \bibinfo{number}{ja} (\bibinfo{year}{2019}).
\newblock
\urldef\tempurl%
\url{https://doi.org/10.1002/ejsp.2615}
\showDOI{\tempurl}
\showeprint{https://onlinelibrary.wiley.com/doi/pdf/10.1002/ejsp.2615}


\bibitem[\protect\citeauthoryear{Conover, Gon{\c{c}}alves, Ratkiewicz,
  Flammini, and Menczer}{Conover et~al\mbox{.}}{2011}]%
        {conover2011predicting}
\bibfield{author}{\bibinfo{person}{Michael~D Conover}, \bibinfo{person}{Bruno
  Gon{\c{c}}alves}, \bibinfo{person}{Jacob Ratkiewicz},
  \bibinfo{person}{Alessandro Flammini}, {and} \bibinfo{person}{Filippo
  Menczer}.} \bibinfo{year}{2011}\natexlab{}.
\newblock \showarticletitle{Predicting the political alignment of twitter
  users}. In \bibinfo{booktitle}{\emph{2011 IEEE third international conference
  on privacy, security, risk and trust and 2011 IEEE third international
  conference on social computing}}. IEEE, \bibinfo{pages}{192--199}.
\newblock


\bibitem[\protect\citeauthoryear{Gayo-Avello}{Gayo-Avello}{2012}]%
        {Gayo-Avello:2012:NYC:2412375.2412758}
\bibfield{author}{\bibinfo{person}{Daniel Gayo-Avello}.}
  \bibinfo{year}{2012}\natexlab{}.
\newblock \showarticletitle{No, You Cannot Predict Elections with Twitter}.
\newblock \bibinfo{journal}{\emph{IEEE Internet Computing}}
  \bibinfo{volume}{16}, \bibinfo{number}{6} (\bibinfo{date}{Nov.}
  \bibinfo{year}{2012}), \bibinfo{pages}{91--94}.
\newblock
\showISSN{1089-7801}
\urldef\tempurl%
\url{https://doi.org/10.1109/MIC.2012.137}
\showDOI{\tempurl}


\bibitem[\protect\citeauthoryear{Gündüz}{Gündüz}{2017}]%
        {TheEffectofSocialMediaonIdentityConstruction}
\bibfield{author}{\bibinfo{person}{Uğur Gündüz}.}
  \bibinfo{year}{2017}\natexlab{}.
\newblock \showarticletitle{The Effect of Social Media on Identity
  Construction}.
\newblock \bibinfo{journal}{\emph{Mediterranean Journal of Social Sciences}}
  \bibinfo{volume}{8}, \bibinfo{number}{5} (\bibinfo{year}{2017}),
  \bibinfo{pages}{85 -- 92}.
\newblock
\urldef\tempurl%
\url{https://content.sciendo.com/view/journals/mjss/8/5/article-p85.xml}
\showURL{%
\tempurl}


\bibitem[\protect\citeauthoryear{Jain and Kumaraguru}{Jain and
  Kumaraguru}{2016}]%
        {Jain:2016:DUC:2888451.2888452}
\bibfield{author}{\bibinfo{person}{Paridhi Jain} {and}
  \bibinfo{person}{Ponnurangam Kumaraguru}.} \bibinfo{year}{2016}\natexlab{}.
\newblock \showarticletitle{On the Dynamics of Username Changing Behavior on
  Twitter}. In \bibinfo{booktitle}{\emph{Proceedings of the 3rd IKDD Conference
  on Data Science, 2016}} \emph{(\bibinfo{series}{CODS '16})}.
  \bibinfo{publisher}{ACM}, \bibinfo{address}{New York, NY, USA}, Article
  \bibinfo{articleno}{6}, \bibinfo{numpages}{6}~pages.
\newblock
\showISBNx{978-1-4503-4217-9}
\urldef\tempurl%
\url{https://doi.org/10.1145/2888451.2888452}
\showDOI{\tempurl}


\bibitem[\protect\citeauthoryear{Johansson, Kyr{\"o}l{\"a}inen, Ginter, Lehti,
  Krizs{\'a}n, and Laippala}{Johansson et~al\mbox{.}}{2018}]%
        {johansson2018opening}
\bibfield{author}{\bibinfo{person}{Marjut Johansson},
  \bibinfo{person}{Aki-Juhani Kyr{\"o}l{\"a}inen}, \bibinfo{person}{Filip
  Ginter}, \bibinfo{person}{Lotta Lehti}, \bibinfo{person}{Attila Krizs{\'a}n},
  {and} \bibinfo{person}{Veronika Laippala}.} \bibinfo{year}{2018}\natexlab{}.
\newblock \showarticletitle{Opening up\# jesuisCharlie anatomy of a Twitter
  discussion with mixed methods}.
\newblock \bibinfo{journal}{\emph{Journal of Pragmatics}}
  \bibinfo{volume}{129} (\bibinfo{year}{2018}), \bibinfo{pages}{90e101}.
\newblock


\bibitem[\protect\citeauthoryear{Katz}{Katz}{1955}]%
        {katzlazarsfeld}
\bibfield{author}{\bibinfo{person}{E Katz}.} \bibinfo{year}{1955}\natexlab{}.
\newblock \bibinfo{title}{y LAZARSFELD, PF (1955): Personal influence: the part
  played by people in the flow of mass communication, Nueva York}.
\newblock
\newblock


\bibitem[\protect\citeauthoryear{Khatua, Cambria, Ghosh, Chaki, and
  Khatua}{Khatua et~al\mbox{.}}{2019}]%
        {Khatua:2019:TSL:3297001.3297057}
\bibfield{author}{\bibinfo{person}{Aparup Khatua}, \bibinfo{person}{Erik
  Cambria}, \bibinfo{person}{Kuntal Ghosh}, \bibinfo{person}{Nabendu Chaki},
  {and} \bibinfo{person}{Apalak Khatua}.} \bibinfo{year}{2019}\natexlab{}.
\newblock \showarticletitle{Tweeting in Support of LGBT?: A Deep Learning
  Approach}. In \bibinfo{booktitle}{\emph{Proceedings of the ACM India Joint
  International Conference on Data Science and Management of Data}}
  \emph{(\bibinfo{series}{CoDS-COMAD '19})}. \bibinfo{publisher}{ACM},
  \bibinfo{address}{New York, NY, USA}, \bibinfo{pages}{342--345}.
\newblock
\showISBNx{978-1-4503-6207-8}
\urldef\tempurl%
\url{https://doi.org/10.1145/3297001.3297057}
\showDOI{\tempurl}


\bibitem[\protect\citeauthoryear{Liu, Lü, Zhang, Tang, and Zhou}{Liu
  et~al\mbox{.}}{2018}]%
        {doi:10.1063/1.5017515}
\bibfield{author}{\bibinfo{person}{Quan-Hui Liu}, \bibinfo{person}{Feng-Mao
  Lü}, \bibinfo{person}{Qian Zhang}, \bibinfo{person}{Ming Tang}, {and}
  \bibinfo{person}{Tao Zhou}.} \bibinfo{year}{2018}\natexlab{}.
\newblock \showarticletitle{Impacts of opinion leaders on social contagions}.
\newblock \bibinfo{journal}{\emph{Chaos: An Interdisciplinary Journal of
  Nonlinear Science}} \bibinfo{volume}{28}, \bibinfo{number}{5}
  (\bibinfo{year}{2018}), \bibinfo{pages}{053103}.
\newblock
\urldef\tempurl%
\url{https://doi.org/10.1063/1.5017515}
\showDOI{\tempurl}
\showeprint{https://doi.org/10.1063/1.5017515}


\bibitem[\protect\citeauthoryear{Mariconti, Onaolapo, Ahmad, Nikiforou, Egele,
  Nikiforakis, and Stringhini}{Mariconti et~al\mbox{.}}{2016}]%
        {Mariconti:2016:WAP:2905760.2905762}
\bibfield{author}{\bibinfo{person}{Enrico Mariconti}, \bibinfo{person}{Jeremiah
  Onaolapo}, \bibinfo{person}{Syed~Sharique Ahmad}, \bibinfo{person}{Nicolas
  Nikiforou}, \bibinfo{person}{Manuel Egele}, \bibinfo{person}{Nick
  Nikiforakis}, {and} \bibinfo{person}{Gianluca Stringhini}.}
  \bibinfo{year}{2016}\natexlab{}.
\newblock \showarticletitle{Why Allowing Profile Name Reuse is a Bad Idea}. In
  \bibinfo{booktitle}{\emph{Proceedings of the 9th European Workshop on System
  Security}} \emph{(\bibinfo{series}{EuroSec '16})}. \bibinfo{publisher}{ACM},
  \bibinfo{address}{New York, NY, USA}, Article \bibinfo{articleno}{3},
  \bibinfo{numpages}{6}~pages.
\newblock
\showISBNx{978-1-4503-4295-7}
\urldef\tempurl%
\url{https://doi.org/10.1145/2905760.2905762}
\showDOI{\tempurl}


\bibitem[\protect\citeauthoryear{Mariconti, Onaolapo, Ahmad, Nikiforou, Egele,
  Nikiforakis, and Stringhini}{Mariconti et~al\mbox{.}}{2017}]%
        {Mariconti:2017:WNU:3038912.3052589}
\bibfield{author}{\bibinfo{person}{Enrico Mariconti}, \bibinfo{person}{Jeremiah
  Onaolapo}, \bibinfo{person}{Syed~Sharique Ahmad}, \bibinfo{person}{Nicolas
  Nikiforou}, \bibinfo{person}{Manuel Egele}, \bibinfo{person}{Nick
  Nikiforakis}, {and} \bibinfo{person}{Gianluca Stringhini}.}
  \bibinfo{year}{2017}\natexlab{}.
\newblock \showarticletitle{What's in a Name?: Understanding Profile Name Reuse
  on Twitter}. In \bibinfo{booktitle}{\emph{Proceedings of the 26th
  International Conference on World Wide Web}} \emph{(\bibinfo{series}{WWW
  '17})}. \bibinfo{publisher}{International World Wide Web Conferences Steering
  Committee}, \bibinfo{address}{Republic and Canton of Geneva, Switzerland},
  \bibinfo{pages}{1161--1170}.
\newblock
\showISBNx{978-1-4503-4913-0}
\urldef\tempurl%
\url{https://doi.org/10.1145/3038912.3052589}
\showDOI{\tempurl}


\bibitem[\protect\citeauthoryear{McCallum}{McCallum}{2002}]%
        {McCallumMALLET}
\bibfield{author}{\bibinfo{person}{Andrew~Kachites McCallum}.}
  \bibinfo{year}{2002}\natexlab{}.
\newblock \bibinfo{title}{MALLET: A Machine Learning for Language Toolkit}.
  (\bibinfo{year}{2002}).
\newblock
\newblock
\shownote{http://mallet.cs.umass.edu.}


\bibitem[\protect\citeauthoryear{Morstatter, Shao, Galstyan, and
  Karunasekera}{Morstatter et~al\mbox{.}}{2018}]%
        {morstatter2018alt}
\bibfield{author}{\bibinfo{person}{Fred Morstatter}, \bibinfo{person}{Yunqiu
  Shao}, \bibinfo{person}{Aram Galstyan}, {and} \bibinfo{person}{Shanika
  Karunasekera}.} \bibinfo{year}{2018}\natexlab{}.
\newblock \showarticletitle{From alt-right to alt-rechts: Twitter analysis of
  the 2017 german federal election}. In \bibinfo{booktitle}{\emph{Companion
  Proceedings of the The Web Conference 2018}}. International World Wide Web
  Conferences Steering Committee, \bibinfo{publisher}{IW3C2},
  \bibinfo{address}{Lyon, France}, \bibinfo{pages}{621--628}.
\newblock


\bibitem[\protect\citeauthoryear{Park}{Park}{2013}]%
        {PARK20131641}
\bibfield{author}{\bibinfo{person}{Chang~Sup Park}.}
  \bibinfo{year}{2013}\natexlab{}.
\newblock \showarticletitle{Does Twitter motivate involvement in politics?
  Tweeting, opinion leadership, and political engagement}.
\newblock \bibinfo{journal}{\emph{Computers in Human Behavior}}
  \bibinfo{volume}{29}, \bibinfo{number}{4} (\bibinfo{year}{2013}),
  \bibinfo{pages}{1641 -- 1648}.
\newblock
\showISSN{0747-5632}
\urldef\tempurl%
\url{https://doi.org/10.1016/j.chb.2013.01.044}
\showDOI{\tempurl}


\bibitem[\protect\citeauthoryear{Raynauld, Richez, and Morris}{Raynauld
  et~al\mbox{.}}{2018}]%
        {doi:10.1080/1369118X.2017.1301522}
\bibfield{author}{\bibinfo{person}{Vincent Raynauld},
  \bibinfo{person}{Emmanuelle Richez}, {and} \bibinfo{person}{Katie~Boudreau
  Morris}.} \bibinfo{year}{2018}\natexlab{}.
\newblock \showarticletitle{Canada is \#IdleNoMore: exploring dynamics of
  Indigenous political and civic protest in the Twitterverse}.
\newblock \bibinfo{journal}{\emph{Information, Communication \& Society}}
  \bibinfo{volume}{21}, \bibinfo{number}{4} (\bibinfo{year}{2018}),
  \bibinfo{pages}{626--642}.
\newblock
\urldef\tempurl%
\url{https://doi.org/10.1080/1369118X.2017.1301522}
\showDOI{\tempurl}
\showeprint{https://doi.org/10.1080/1369118X.2017.1301522}


\bibitem[\protect\citeauthoryear{Ryan~Wesslen}{Ryan~Wesslen}{2018}]%
        {DBLP:conf/icwsm/WesslenNEGLJS18}
\bibfield{author}{\bibinfo{person}{Omar Eltayeby Tiffany Gallicano Sara Levens
  Min Jiang Samira~Shaikh Ryan~Wesslen, Sagar~Nandu}.}
  \bibinfo{year}{2018}\natexlab{}.
\newblock \showarticletitle{Bumper Stickers on the Twitter Highway: Analyzing
  the Speed and Substance of Profile Changes}. In
  \bibinfo{booktitle}{\emph{Proceedings of the Twelfth International Conference
  on Web and Social Media, {ICWSM} 2018, Stanford, California, USA, June 25-28,
  2018.}} \bibinfo{publisher}{{AAAI} Press}, \bibinfo{address}{California,
  USA}, \bibinfo{pages}{696--699}.
\newblock
\urldef\tempurl%
\url{https://aaai.org/ocs/index.php/ICWSM/ICWSM18/paper/view/17834}
\showURL{%
\tempurl}


\bibitem[\protect\citeauthoryear{Schwämmlein and Wodzicki}{Schwämmlein and
  Wodzicki}{2012}]%
        {10.1111/j.1083-6101.2012.01582.x}
\bibfield{author}{\bibinfo{person}{Eva Schwämmlein} {and}
  \bibinfo{person}{Katrin Wodzicki}.} \bibinfo{year}{2012}\natexlab{}.
\newblock \showarticletitle{What to Tell About Me? Self-Presentation in Online
  Communities}.
\newblock \bibinfo{journal}{\emph{Journal of Computer-Mediated Communication}}
  \bibinfo{volume}{17}, \bibinfo{number}{4} (\bibinfo{date}{07}
  \bibinfo{year}{2012}), \bibinfo{pages}{387--407}.
\newblock
\showISSN{1083-6101}
\urldef\tempurl%
\url{https://doi.org/10.1111/j.1083-6101.2012.01582.x}
\showDOI{\tempurl}
\showeprint{http://oup.prod.sis.lan/jcmc/article-pdf/17/4/387/22317340/jjcmcom0387.pdf}


\bibitem[\protect\citeauthoryear{Times}{Times}{2019}]%
        {chowkidar}
\bibfield{author}{\bibinfo{person}{The~Economic Times}.}
  \bibinfo{year}{2019}\natexlab{}.
\newblock \bibinfo{title}{'Chowkidar Narendra Modi': PM changes Twitter handle
  name to counter Rahul Gandhi's chor jibe}.
\newblock
\newblock
\urldef\tempurl%
\url{https://economictimes.indiatimes.com/news/politics-and-nation/chowkidar-narendra-modi-pm-changes-twitter-handle-name-to-counter-rahul-gandhis-chor-jibe/articleshow/68448053.cms}
\showURL{%
\tempurl}


\bibitem[\protect\citeauthoryear{Twitter}{Twitter}{2012}]%
        {twitterVerified}
\bibfield{author}{\bibinfo{person}{Twitter}.} \bibinfo{year}{2012}\natexlab{}.
\newblock \bibinfo{title}{"Please note: changing your username will result in
  losing your badge."}.
\newblock
\newblock
\urldef\tempurl%
\url{https://twitter.com/verified/status/225985823798091776}
\showURL{%
\tempurl}


\bibitem[\protect\citeauthoryear{Varol, Ferrara, Ogan, Menczer, and
  Flammini}{Varol et~al\mbox{.}}{2014}]%
        {Varol:2014:EOU:2615569.2615699}
\bibfield{author}{\bibinfo{person}{Onur Varol}, \bibinfo{person}{Emilio
  Ferrara}, \bibinfo{person}{Christine~L. Ogan}, \bibinfo{person}{Filippo
  Menczer}, {and} \bibinfo{person}{Alessandro Flammini}.}
  \bibinfo{year}{2014}\natexlab{}.
\newblock \showarticletitle{Evolution of Online User Behavior During a Social
  Upheaval}. In \bibinfo{booktitle}{\emph{Proceedings of the 2014 ACM
  Conference on Web Science}} \emph{(\bibinfo{series}{WebSci '14})}.
  \bibinfo{publisher}{ACM}, \bibinfo{address}{New York, NY, USA},
  \bibinfo{pages}{81--90}.
\newblock
\showISBNx{978-1-4503-2622-3}
\urldef\tempurl%
\url{https://doi.org/10.1145/2615569.2615699}
\showDOI{\tempurl}


\end{thebibliography}










\end{document}